\documentclass[letterpaper, 10 pt, conference]{ieeeconf}  % Comment this line out if you need a4paper

\IEEEoverridecommandlockouts                              % This command is only needed if 
\usepackage{algorithm, algpseudocode}
\usepackage{amsmath}
\usepackage{graphicx}
\usepackage{url}
\usepackage{amssymb}
\usepackage{booktabs}
\usepackage{makecell}
\usepackage{xcolor}
\usepackage{hyperref}

\title{\LARGE Can I join you? Extended Social Force Based Model with Predictive Control and Barrier Functions for Humanoid Group Repositioning Amidst Obstacles}
\author{Chuxuan Zhang\textsuperscript{1}, Niloofar Jahanshahi\textsuperscript{1}, Yasaman Etesam\textsuperscript{1}, Mo Chen\textsuperscript{1}, Angelica Lim\textsuperscript{1}% <-this % stops a space
\thanks{$^{1}$Simon Fraser University, Computing Science,
        Burnaby, BC, Canada
        Contact: {\tt\small chuxuan\_zhang@sfu.ca}}%
}

\begin{document}

\maketitle
% \thispagestyle{empty}
% \pagestyle{empty}

% abstract + title + contributions
% IMLE social force
% explanable and probabilistic model that capture realistic

% related work (how each of the contribution is important and not ready)
% social force model 
% group joning

\begin{abstract}
As social and service robots increasingly share interaction spaces in public, they may form groups while coordinating tasks or interacting with humans. Initial group formation alone is insufficient in dynamic settings, because it can be disrupted when a person or another agent joins, leaves, or passes through the group. Although group dynamics and repositioning have been widely studied in swarm robotics and virtual-agent settings, little attention has been given to socially appropriate repositioning behaviors in groups of humanoid robots. Existing social-force-based approaches provide a useful foundation for group behavior, but they have limited obstacle handling, and can produce unsafe behavior as environmental complexity increases. We propose a novel framework that modulates traditional Social Force, guided by Model Predictive Control (MPC) to plan ahead, bounded by Control Barrier Functions (CBF) to increase safety for robot groups repositioning behaviors. This approach aims to produce safer and socially aware group repositioning behaviors under various group formation disturbance. We evaluate the method in simulation through an ablation study and an online user study. The results show that our proposed method can produce smoother and safer repositioning behaviors in complex environments while preserving social appropriateness.

\end{abstract}

\section{Introduction}
Group repositioning is common in everyday social interactions. We may need to join or leave colleagues who are chatting in an office or navigate through crowded poster sessions at a conference. Similar situations can also arise in human--robot group interactions, for example when a person enters an elevator occupied by a group of delivery robots or approaches a group of tour-guide robots while traveling. In these situations, existing group members may need to reposition themselves to acknowledge the newcomer, provide access to the shared interaction space, or allow the person to pass through comfortably.
When existing group members fail to acknowledge and accommodate a newcomer’s joining or leaving behavior, additional time and effort may be required for the newcomer to engage with the group, while the remaining members may struggle to maintain an appropriate group formation. A lack of adaptive proxemic behavior may also be perceived as passive-aggressive toward a passerby and make navigation through the shared space uncomfortable. As humanoid robots become more widely deployed in real-world environments, improving the repositioning behavior of existing robot group members is important for enabling smooth and natural group interactions.

\begin{figure}[t]
    \centering
    \includegraphics[width=\linewidth]{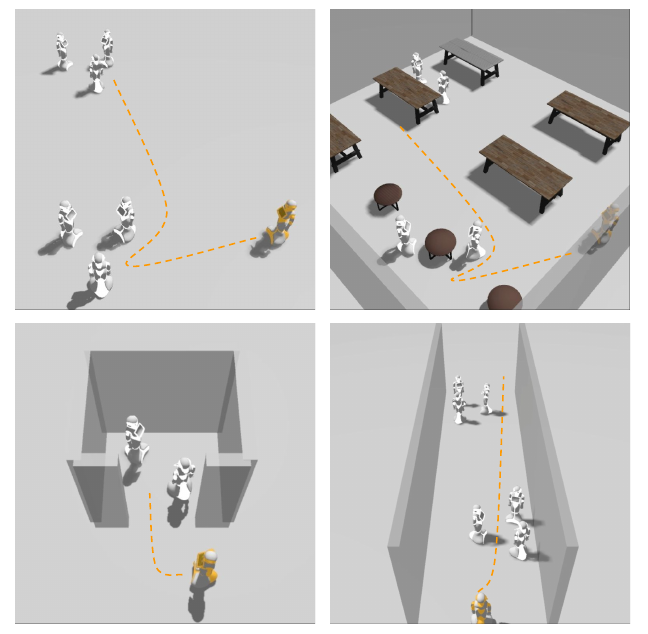}
    \caption{\textbf{Snapshots of four Gazebo Fortress layouts: empty space, crowded room, elevator, and hallway.} The white robot groups reposition themselves as the yellow robot approaches along the dashed trajectory.}
    \label{fig:study_video_snapshot}
\end{figure}

The Social Force Field Model (SFFM) \cite{pedica2008social} was proposed in 2008 for free-standing conversational virtual agents in obstacle-free environments. SFFM succeeds in static group formation contexts, suggesting it's better suited to maintaining a formation once established than to producing naturalistic, adaptive dynamics \cite{gao2019learning}. Nevertheless, it still lacks the capability to maintain appropriate group formation during real-world interaction. Gao et al. \cite{gao2019learning} applied SFFM in robot group joining study, demonstrating its potential to translate from virtual agents setting to physical robot deployment. However, transferring SFFM from virtual agents to physical robot groups introduces additional safety concerns. Robots deployed in real environments rarely operate in obstacle-free spaces, and collisions involving physical robots can be expensive. In this study, we propose an Extended Social Force Field Model (ESFFM) that not only produces initial static group formation, but also adapts to disturbances in group formation.

When a social-force-inspired controller is used alone for group repositioning, force cancellation cannot be prevented, and complex environments can increase the likelihood of unsafe proxemic behavior. Therefore, we introduce receding-horizon prediction to optimize group repositioning behavior over a longer interaction horizon, together with barrier functions that provide an additional safety mechanism. While barrier functions provide an additional safety filter, they do not themselves encode social context. This motivates our framework, which combines social-force-inspired optimization with barrier-function-based safety constraints to produce safer and more natural group dynamics for social humanoids.

In this paper, we focus on modeling the adaptive behavior of existing robot group members when the group formation is disturbed by joining, leaving, or passing-by behaviors. Compared with prior work on virtual-agent group dynamics, we translate group-repositioning behavior to social robot settings and evaluate it across a variety of environments. The major contributions of this paper are summarized as follows:
\begin{enumerate}

    \item An obstacle-aware Extended Social Force Field Model (ESFFM) for adaptive robot-group repositioning under interaction disturbances.
    \item A receding-horizon controller that optimizes ESFFM behavior over a finite horizon to reduce unnecessary motion.
    \item An ESFFM-based predictive controller with barrier-function safety filtering that enables safer robot-group repositioning across varied environments while preserving social appropriateness.

\end{enumerate}

\section{Related Work}
\subsection{F-formation and Group Reformation}
Kendon’s concept of F-formation \cite{kendon1990conducting} provides a useful lens for analyzing group dynamics. An F-formation describes the spatial patterns formed during face-to-face interactions. Human F-formation is often studied in robot social navigation \cite{yoo2026did} or group joining \cite{barua2020let, narayanan2015equitably, nigro2025social}, and with virtual avatars \cite{kum2025joining}.

% When 2 or more people form a face-to-face interaction group, they share a transactional segment (o-space) that is equally accessible and exclusive to the group members. The p-space is the area where all group member occupy and surrounding the o-space. The outer space is the r-space surrounding the group, can be extended indefinitely to distinguish the rest of the world scene with the group. 

The Social Force Model (SFM) \cite{helbing1995social}, originally proposed by Helbing to simulate pedestrian behavior, has been widely studied and incorporated into social-navigation simulation platforms for modeling human motion in robot navigation tasks \cite{tsoi2022sean, perez2023hunavsim, kastner2024arena}. Beyond modeling individual humans in crowd navigation, this physics-based model has also been extended to group navigation \cite{moussaid2010walking, huang2018social}. Pedica and Vilhjálmsson \cite{pedica2008social} applied force-field rules to maintain F-formations among conversational agents by introducing equality and cohesion forces within the group. Kuzuoka et al. \cite{kuzuoka2010reconfiguring} studied how a robot’s rotational behavior could lead people to reconfigure their F-formation arrangement. Yang et al. \cite{yang2017study} studied robot joining behaviors that can either disturb or maintain the F-formation of a human group. Despite the aforementioned work on maintaining F-formations in virtual agent and human groups, group repositioning behaviors among social humanoids remain understudied.

% \begin{figure}
%     \centering
%     \includegraphics[width=0.8\linewidth]{ieeeconf/figure/F-formation.pdf}
%     \caption{Structure of F-formation. Black circles denote social agents. The o-space is the shared transactional region, the p-space is occupied by group members surrounding the o-space, and the r-space serves as a buffer separating the group from its environment.}
%     \label{fig: F-formation}
% \end{figure}

\subsection{Model Predictive Control with Social Objectives and Spatial constraints}
Socially aware navigation has established the receding-horizon formulation as an effective way for achieving social objectives. The Social Force Model itself has been integrated with model predictive control (MPC) for socially aware navigation in mixed traffic \cite{zhang2023study}. Trepella et al. \cite{trepella2026navigating} applied social cost functions to guide MPC optimization toward human-compliant behaviors with simulated humans. Sacco et al. \cite{sacco2024novel} combined the predictive capability of MPC with the reactivity of the Social Force Model to model pedestrian dynamics. Fiaschè et al. \cite{fiasche2023towards} proposed the universal social force model as a prediction model in a parameterized MPC framework. While MPC can enforce hard constraints such as minimal distance from obstacles or actuation limits, these constraints are imposed only at the sampled points of a finite horizon, and can not provide a guarantee of persistent safety.

\subsection{Control Barrier Functions in Social Navigation}
In contrast, control barrier functions (CBF) constrain the rate of change of a safety margin to render the safe set forward invariant. Zeng et al. \cite{zeng2021safety} proposed discrete-time CBF formulation imposed as constraints across the MPC horizon for ego-vehicle driving.  Combined MPC-CBF frameworks have also been used to ensure safe and deadlock-free navigation for decentralized multi-robot systems in constrained environments \cite{chandra2025deadlock}, dynamic environments \cite{magalhaes2025distributed}, and human-following robot tasks \cite{li2025mpc}.

While effective for safety, CBFs are primarily geometric and do not encode social preferences, motivating hybrid approaches with socially informed controllers. Hence, we propose a group repositioning framework for social humanoids that uses receding-horizon predictive planning with a social-force-inspired nominal controller, further guarded by a CBF to avoid collisions.

\section{Problem Definition}
\label{sec: problem definition}
This study focuses on autonomous group reformation behaviors when disturbed by a teleoperated robot's proxemic behavior. The task is to compute the next state of a target robot $s_i(t+1), i \in[1,\ldots,N]$ given the current state of all robots $s_0(t), s_1(t), \ldots, s_N(t)$, 
% \NJ{$s_1$ missing} 
and $N+1$ is the total number of robots. We set $\mathrm{robot}_0$ as the teleoperated robot for this study, and the rest $N$ robots are autonomous target robots driven by the group controller. 

The teleoperated robot serves as an externally controlled agent whose motion is independent of the proposed framework. Accordingly, this robot
is treated as an external disturbance to the robot groups rather
than as an agent whose motion is generated by the group
controller. At each control step, only its currently observed
state is available to the group controller. It can be viewed as a proxy for a human or other agent interacting with the robot groups in real-world settings.

\textbf{Robot State Representation} Each robot state $s_{k}(t)$ contains: 1) spatial position $[x,y]$, 2) orientation $\theta$, 3) velocity $[v_x, v_y]$ in the global frame, 4) group number $gn$ an integer that encodes its group membership, and 5) identity, a binary code ${0/1}$ indicating if the robot is teleoperated or autonomous.

\section{Methods}
% Add the purpose of our approach, describe inputs and outputs
\begin{figure*}[t]
    \centering
    \includegraphics[width=\linewidth]{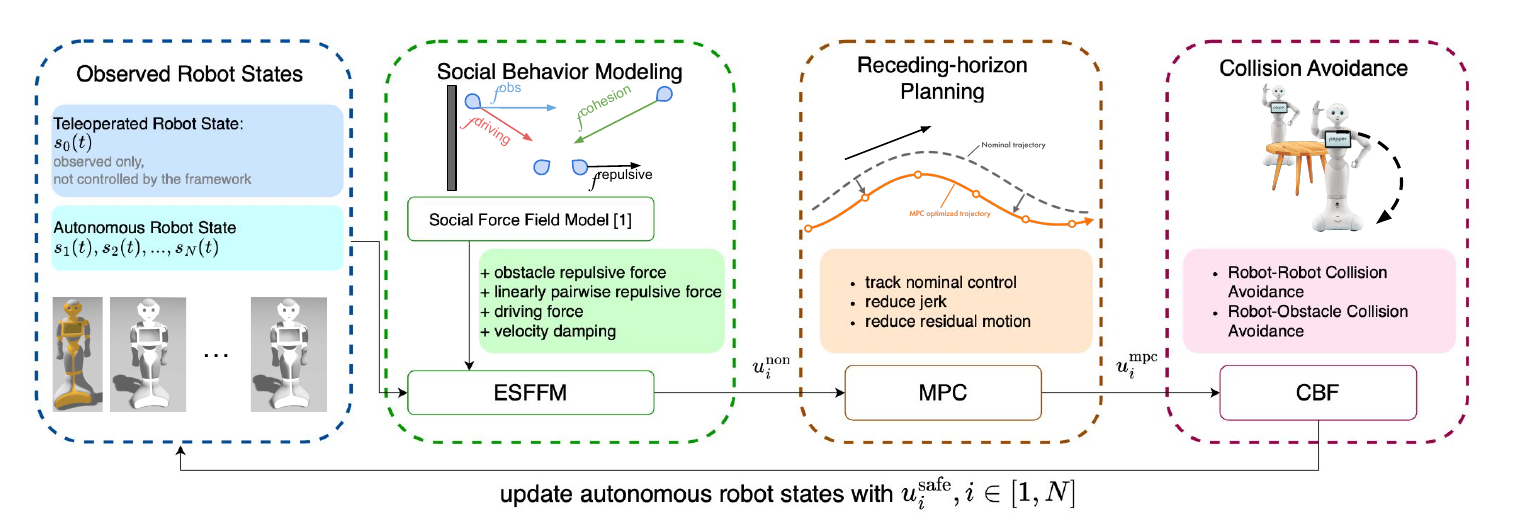}
    \caption{
   \textbf{Overview of our proposed robot-group repositioning framework.} ESFFM generates nominal accelerations from observed robot and environment states, which are refined by MPC and filtered by CBF. Safe accelerations are applied only to autonomous robots, and the teleoperated robot acts as an external disturbance.}
    \label{fig:flowchart}
\end{figure*}

The goal of this study is to model the naturalistic and safe repositioning behaviors for robot groups in various social environment settings with obstacles. In this section, we will start with the original SFFM, followed by our extended SFFM that adapts to the study environment setup, the MPC optimizer and the CBF filter.

\subsection{Social Force Field Model}
\label{sec: SFFM}
SFFM was originally designed for group dynamics modeling in virtual-avatar settings. The original model is composed of three forces: repulsive force, equality force and cohesion force that influence the positioning and orientation of each group members. The forces are:
% requires \usepackage{amsmath}
\begin{align}
F^\mathrm{repulsive} &= -\left(\Delta_p - d_{min}\right)^2 \frac{R}{\|R\|} \label{eq:frep}\\
F^\mathrm{equality}  &= \left(1 - \frac{m}{|c - r|}\right)(c - r) \label{eq:feq}\\
F^\mathrm{cohesion}  &= \alpha \left(1 - \frac{s}{\|o - r\|}\right)(o - r) \label{eq:fcoh}
\end{align}

\noindent with the associated reorientation terms

\begin{align}
D_{equality} &= \sum_{k}^{N_s}\left(r_k - r\right), &
D_{cohesion} &= \sum_{k}^{N_a}\left(r_k - r\right)
\label{eq:dterms}
\end{align}

\noindent where $r$ is the avatar position, $o$ is the conversation center, $s$ is the o-space
radius, $\Delta_p$ is the personal distance, and $N_p$, $N_s$, $N_a$ are the number of
individuals in the avatar's personal, social and public space respectively. The
aggregate $R$, group centroid $c$, mean member distance $m$ and cohesion
scaling $\alpha$ are given by Eqs.~\eqref{eq:R} --\eqref{eq:alpha}. And the derived quantities are:

\begin{align}
R      &= \sum_{k}^{N_p}\left(r_k - r\right) \label{eq:R}\\
c      &= \frac{1}{N_s + 1}\left(r + \sum_{k}^{N_s} r_k\right) \label{eq:c}
\end{align}

\begin{align}
m      &= \frac{\|r - c\|}{N_s + 1} \sum_{k}^{N_s} \|r_k - c\| \label{eq:m}\\
\alpha &= \frac{N_a}{N_s + 1} \label{eq:alpha}
\end{align}
% \NJ{$c$ equation sums over $i$ but uses $r_k$}

\subsection{Extended Social Force Field Model}
\label{sec: ESFFM}
Despite its effectiveness, SFFM has several limitations when applied to physical robot groups. First, SFFM was originally designed for free-standing conversational groups in open, obstacle-free environments. Translating this setting from virtual-agent scenarios to real-world robot deployment is challenging because physical robots rarely operate in such unconstrained spaces. Second, SFFM lacks a global anchor for each conversational group. A passing robot that displaces one group member can consequently cause the entire group to drift away from its initial gathering position, with no mechanism for recovering that position after the disturbance. Third, the repulsive force formulation can be trapped in a local minimum. 
For example, when one robot is 0.2 m ahead and three robots are 1.0 m behind, the centroid of the neighboring robots lies behind the target robot. The resulting repulsive force will drive the robot forward toward its nearest neighbor, while the force magnitude increases as $d_{\min}$ decreases, potentially pinning the robot against the agent it should be avoiding. Fourth, Helbing's original SFM \cite{helbing1995social} includes damping behavior through relaxation time parameter in the self-driving force. In contrast, SFFM does not specify an explicit damping mechanism, which can lead to repeated closing-and-opening oscillations within the group when the formation is not initialized near equilibrium or kept balanced after being disturbed. Finally, Pedica and Vilhjálmsson \cite{pedica2008social} described the outputs of SFFM as motivational forces, without specifying an explicit mapping from these forces to motor control. To establish a comparable baseline for physical robot deployment, we first introduce the Extended Social Force Field Model (ESFFM) to address these limitations and then define the corresponding system dynamics.

The key extensions introduced in ESFFM are:
\begin{enumerate}
\item a \textbf{linear obstacle-repulsive force} Eq. \eqref{eq:obs} based on the distance between each autonomous robot and nearby obstacles;
\item a \textbf{quadratic driving force}  Eq.\eqref{eq:driving} that attracts group members toward a global group center, mitigating group drift and limiting disruptive displacement of the group as a whole;
\item a \textbf{pairwise robot--robot repulsive force} Eq.\eqref{eq:revised_repulsion} with a linear force magnitude, replacing the original quadratic formulation to reduce repulsive-force overshoot and improve stability in interpersonal spacing; and
\item a \textbf{linear velocity-decay rate $\gamma$} Eq.\eqref{eq:decay} that stabilizes motion, reduces oscillations, and produces more naturalistic repositioning behavior.
\end{enumerate}

The forces acting on a single autonomous $\mathrm{robot}_i$ are thus:

\begin{align}
    F^\mathrm{obs}& =
        \sum_{o \in \mathcal{O}}
        \max\left(0,\; d_{\mathrm{th}} + r_{\mathrm{inflate}} - d_{io}\right)
        \mathbf{n}_{io} \label{eq:obs}\\
    F^\mathrm{driving} &=
        \left\|\mathbf{c}_{i}-r_{i}\right\|^{2}
        \frac{\mathbf{c}_{i}-r_{i}}
        {\left\|\mathbf{c}_{i}-r_{i}\right\|+\varepsilon} \label{eq:driving} \\
    F^\mathrm{repulsion}& = -\sum_{k}^{N_p} \left( \Delta_{i,k} - d_{i,k} \right) \frac{r_k - r_i}{\left\| r_k - r_i \right\|}
    \label{eq:revised_repulsion}
\end{align}

\noindent where $d_{\mathrm{th}}$ denotes the activation-distance threshold of the obstacle-repulsive force, $r_{\mathrm{inflate}}$ is half the width of the robot, and $d_{io}$ is the distance between $\mathrm{robot}_i$ and obstacle $o$. The unit vector $\mathbf{n}_{io}$ points away from obstacle $o$. The vector $\mathbf{c}_i$ denotes the global center of the group containing $\mathrm{robot}_i$.
For the pairwise robot--robot repulsion, $r_k-r_i$ denotes the relative position vector from the target $\mathrm{robot}_i$ to the neighbour $\mathrm{robot}_k$, and $d_{i,k}$ is the Euclidean distance between them. The desired personal distance between the target $\mathrm{robot}_i$ and $\mathrm{robot}_k$ is defined as
$\Delta_{i,k}=\frac{1}{2}\left(\Delta_p+\Delta_{i}\right)$, i.e., the mean of their preferred personal distances. We adopt the original SFFM cohesion and equality forces equations for ESFFM.

We interpret the social force terms as acceleration commands for a self-driven robot rather than as mechanical forces. This interpretation is consistent with Helbing's original characterization of social forces as a quantification of internal motivation rather than physical interaction. We define the dynamics of autonomous robot\ $i$ as:

\begin{subequations}
\label{eq: dynamics}
\begin{align}
\mathbf p_{i,n+1}
&= \mathbf p_{i,n}
+ \Delta t \mathbf v_{i,n}
+ \tfrac{1}{2}\Delta t^2\mathbf u_{i,n}, \\
\mathbf v_{i,n+1}
&= \rho\left(\mathbf v_{i,n}
+ \Delta t \mathbf u_{i,n}\right),\\
\rho &= e^{-\gamma\Delta t},
 \label{eq:decay}
\end{align}    
\end{subequations}

\noindent where $n$ denotes the discrete timestep, $\Delta t$ is the prediction interval, and $\gamma$ represents velocity damping in the motion model. Let $\mathbf p_i\in\mathbb R^2$ and $\mathbf v_i\in\mathbb R^2$ denote the planar position and velocity of $\mathrm{robot}_i$, respectively, and let $\mathbf u_i\in\mathbb R^2$ denote its acceleration command.
Following the later improvement to SFFM proposed by Pedica and Vilhjálmsson \cite{pedica2010spontaneous}, we also adopt constant-velocity extrapolation to predict the future positions of neighboring robots and compute the corresponding reactive social forces in the baseline ESFFM.

To combine the different social-force components into a single nominal control input, each force is scaled by an individual robot weighting factor $K^f_i$ where $f$ indexes the force type, allowing different robots to exhibit distinct proxemic preferences. For simplicity, each robot remains oriented toward the dynamically updated center of its assigned group, computed from all group-member positions, rather than using the original SFFM equality- and cohesion-force directions. (Eq. \eqref{eq:dterms}). This allows the model to focus on repositioning rather than reorientating as the primary adaptive response.
    
\subsection{Model Predictive Control}
Building on the ESFFM formulation, we introduce a model predictive planner that evaluates social interactions over a finite prediction horizon. While ESFFM only provides a behavioral preference for the current state based on social cues, optimizing over a longer horizon can smooth trajectory and reduce residual motion while guided by the social context.

Let $\mathcal A$ denote the set of autonomous robots. At each control update, the MPC optimizes the acceleration inputs $\mathbf u_{i,n}$ over a horizon of $H$ prediction steps. Let $\mathbf F_i^{\mathrm{esffm}}(\mathbf P_n,\mathbf V_n)$ denote the ESFFM acceleration evaluated at prediction step $n$. This force is recomputed at every stage of the prediction horizon. The optimization objective is

\begin{equation}
\begin{split}
J=\sum_{n=0}^{H-1}\sum_{i\in\mathcal A}\Big[
&\tfrac{q_n}{2}\big\|\mathbf u_{i,n}-\mathbf F_i^{\mathrm{esffm}}\big\|^2
+w_{\Delta u}\big\|\mathbf u_{i,n}-\mathbf u_{i,n-1}\big\|^2\\
&+w_s\big[\|\mathbf v_{i,n+1}\|-v_{\max}\big]_+^2
+w_v\,\eta_n\big\|\mathbf v_{i,n+1}\big\|^2\\
&+w_e\big\|\mathbf F_{i,n}^{\mathrm{eq}}\big\|^2\Big].
\end{split}
\end{equation}

subject to the prediction dynamics in \eqref{eq: dynamics} and the acceleration constraint $||\mathbf u_{i,n}||\le a_{\max}$. Here, $\eta_n=n/\max(H-1,1)$, and $\mathbf F_{i,n}^{\mathrm{eq}}$ denotes the equality-force component defined by ESFFM. The tracking term weighted by $q_n$ preserves the motivational behavior generated by ESFFM. The acceleration-difference term,
$w_{\Delta u}||\mathbf u_{i,n}-\mathbf u_{i,n-1}||^2$,
penalizes abrupt changes in the control input and thereby reduces jerk. The soft speed-limit term, $w_s[||\mathbf v_{i,n+1}||-v_{\max}]+^2$,
penalizes predicted speeds above $v{\max}$. The stage-dependent velocity penalty,
$w_v\eta_n||\mathbf v_{i,n+1}||^2$,
increases along the prediction horizon and encourages the robots to reduce residual motion toward the end of the horizon. The equality-force penalty discourages predicted formation states that require large equality corrections. Together, the velocity and equality-force penalties encourage faster settling of the group reformation behavior. All of these objective terms express soft preferences rather than guarantees of convergence. The nonlinear objective is differentiated through the prediction rollout and approximately minimized using a fixed number of projected Adam iterations, with gradient clipping and a horizon length of $H=20$.

\subsection{Control Barrier Functions}
To further improve safety, a control barrier function (CBF) filter is placed directly after ESFFM or after MPC to modify the nominal acceleration according to physical separation and control constraints. 

For the stacked planar state $\mathbf x$, the filter uses the control-affine system:

\begin{equation}
\label{eq:cbf_dynamics} \dot{\mathbf p}_i=\mathbf v_i,\qquad \dot{\mathbf v}_i=-\gamma\mathbf v_i+\mathbf u_i, \qquad \dot{\mathbf x}=f(\mathbf x)+g(\mathbf x)\mathbf u. 
\end{equation}

We formulate the safety filter using continuous-time high-order control barrier functions under damped double-integrator dynamics. The resulting control barrier function quadratic program (CBF-QP) is evaluated at discrete control instants with a fixed sampling period, and each computed acceleration is held until the next update. Thus, the controller has a sampled-data implementation of a continuous-time CBF rather than a discrete-time CBF.

% \NJ{The CBF is based on the continuous-time model in \eqref{eq:cbf_dynamics}, while the controller above uses the discrete dynamics in \eqref{eq: dynamics}. Are we basically treating the discrete model as an approximation of the continuous one here? I also think we should be a bit careful with saying that the CBF gives a strict collision-free guarantee, since it is applied at discrete updates and the teleoperated robot is externally controlled. The filter can react through the autonomous robots, but it cannot control the teleoperated robot if it moves directly toward them. Maybe better to consistently describe it as a safety filter or safety intervention.}\shay{revised}

% The damping coefficient models velocity attenuation in the motion controller. It is part of the assumed dynamics rather than an additional damping objective imposed by the safety filter.

For $\mathrm{robot}_i$ and $\mathrm{robot}_k$, physical separation is represented by
\begin{equation}
\label{eq:robot_robot_cbf}
h_{ik} = \|\mathbf p_i-\mathbf p_k\|^2-d_{\mathrm{safe}}^2.
\end{equation}

where $d_{\mathrm{safe}}$ is the required center-to-center distance derived from the robot footprint. For a circular obstacle with center $\mathbf c_o$ and radius $R_o$,

$$ h_{io}^{\mathrm{circ}}=|\mathbf p_i-\mathbf c_o|^2-(R_o+r^\mathrm{safe})^2, $$

where $r^\mathrm{safe}$ is the robot safety radius. Rectangular obstacles are represented using an inflated fourth-order superellipse. With obstacle-aligned coordinates $\mathbf z=R(\theta_o)^\top(\mathbf p_i-\mathbf c_o)$,

$$
 h_{io}^{\mathrm{box}} = \left(\frac{z_x}{a_o}\right)^4+ \left(\frac{z_y}{b_o}\right)^4-1, 
 $$
 
$$ 
 \quad a_o =2^{1/4}\frac{\ell_o}{2}+r,\quad b_o=2^{1/4}\frac{w_o}{2}+r. 
$$

Here, $\ell_o$, $w_o$, and $\theta_o$ are obstacle length, width, and orientation respectively. The inflated axes provide clearance around the rectangular footprint. For these geometric barriers, $h\ge0$ denotes the admissible side of the boundary. A separate speed barrier is defined as

$$ h_i^v=v_{\max}^2-|\mathbf v_i|^2. $$

Because acceleration affects geometric separation through its second time derivative, the position barriers are converted to relative-degree-one barriers:

\begin{equation}
\label{eq:hocbf_condition}
\psi_1=\dot h+\alpha_1 h,\qquad
\dot \psi_1+\alpha_2 \psi_1\ge0.
\end{equation}
with initial conditions satisfying $h\geq0$ and $\psi_1\geq0$, and positive gains $\alpha_1$ and $\alpha_2$. Equivalently,

$$ \ddot h+(\alpha_1+\alpha_2)\dot h+\alpha_1\alpha_2 h\ge0. $$

These expressions make the barrier inequality affine in acceleration. The speed barrier already has relative degree one and uses $\dot h_i^v+\alpha_v h_i^v\ge0$ directly.

The filtered acceleration is obtained from

 \begin{equation} 
 \label{eq:cbf_qp}
 \begin{split}
 &\mathbf u^{\mathrm{safe}}=\arg\min_{\mathbf u} \frac12||\mathbf u-\mathbf u^{\mathrm{nom}}||^2\ \text{subject to};\\
 & L_f\psi_1+L_g\psi_1\mathbf u+\alpha_2 \psi_1\ge0,\quad \forall h,\\ &L_fh_i^v+L_gh_i^v\mathbf u+\alpha_v h_i^v\ge0,\quad \forall i,
 \\ &-\bar u\le u_{i,x},u_{i,y}\le\bar u. 
 \end{split}
 \end{equation}

The objective selects the feasible acceleration closest to the nominal command generated either directly by ESFFM or by MPC. The Lie derivatives $L_f$ and $L_g$ describe how each barrier function evolves under the uncontrolled dynamics and control input, respectively. The control bound $\bar{u}$ limits the QP actuation. Barrier constraints are enforced as hard constraints without relaxation variables. When obstacle selection is enabled, each robot considers a fixed number of nearest static obstacle footprints. If the solver reports a failure or returns an invalid output, zero-velocity commands are sent to the autonomous robots.

In the current implementation, all robots are included in the QP dynamics and pairwise safety constraints. The control input $u_0$ of the teleoperated robot is treated as an external input rather than a decision variable, and corrective actions are applied only to the autonomous robots. At each CBF update, the latest observed position of the teleoperated robot is used and held fixed until the next control step, at which point its position is updated again. This treatment avoids introducing an unreliable velocity prediction into the higher-order barrier constraints and improves optimization feasibility. 
% \NJ{Are we currently setting the teleoperated robot's velocity to zero in the CBF state, as stated here? Since $\dot h_{i0}$ for the pairwise barrier in Eq \eqref{eq:robot_robot_cbf} depends on the relative velocity $\mathbf v_i-\mathbf v_0$ and enters the HOCBF condition in Eq \eqref{eq:hocbf_condition}, are we instead using the teleoperated robots measured or commanded velocity in the actual implementation?} 
The final control output satisfies the hard actuation constraints.
% \NJ{To me, this part seems to be written a bit inconsistently. Saying that the final CBF output satisfies the hard actuation   constraints (written above) seems to contradict the below statement that the "post-CBF hard actuation limits"  may violate the dynamics assumed by the optimization.}
Accordingly, the implemented CBF filter should be interpreted as a safety intervention rather than a guarantee of collision-free behavior. The motion of the externally controlled robot may violate the dynamics assumed by the optimization.
% \NJ{I think this is the right way to describe the CBF here. We should keep this wording consistent throughout the paper and avoid saying that the CBF guarantees safety, since the teleoperated robot is externally controlled and can move in a way that the safety filter cannot enforce.}

\section{User Study}
\label{sec:user_study}
We conducted an online user study to evaluate perceived interaction quality of robot-group repositioning produced by our proposed framework (ESFFM+MPC+CBF). Because the original SFFM lacks basic obstacle-aware behavior, we use ESFFM as the baseline for a fairer comparison, allowing the study to focus on the perceptual effects of predictive optimization and safety filtering rather than SFFM’s limitations in obstacle-rich environments.

\subsection{Study Design}
Participants were instructed to watch pairs of videos and compare robot group repositioning behaviors caused by teleoperated robot's disturbance using a set of rating scales.

The study used a within-subjects design with two factors: the physical layout of the scene (Fig. \ref{fig:study_video_snapshot}) and the number of agents present in each scene. The scene layout had four levels: empty space, crowded room, elevator, and hallway. The number of robots had two levels: 5 vs. 7 agents for the empty space, crowded room, and hallway conditions, and 2 vs. 3 agents for the elevator condition. These robot counts included the externally controlled robot. 

This resulted in 8 conditions, each presented as one pair of videos. Participants evaluated two videos of one pair vertically displayed on the same page. To control for order effects, the order of the eight video pairs was randomized, and the top-bottom display order of the two videos within each pair was also randomized. Within each condition, the video pair used the same pre-recorded trajectory for the teleoperated robot, ensuring that differences between the videos are from the group controllers rather than from the disturbances. Participants were instructed to focus on the repositioning behavior of the robot groups rather than the teleoperated robot.
% Before beginning the main task, participants completed a comprehension check asking which robot behaviors they should observe. Participants who answered incorrectly were returned to the instructions and asked to try again. The main task continued only after the comprehension check was answered correctly.

\subsection{Stimuli}
Each video lasted approximately 10--30 seconds, depending on the complexity of the repositioning behavior and the scene. The participants could replay them as many times as needed while answering the questions. For the empty space and crowded room conditions, as well as the hallway condition with 5 agents, the joining robot completed two joining tasks in sequence. For the hallway condition with 7 agents and the elevator conditions, the joining robot completed one joining task.

\subsection{Measures}
For each video pair, participants answered 8 questions. Eight questions used a 7-point comparative Likert scale and were derived from the evaluation used by Pedica et al. \cite{pedica2010avatars} for SFFM. These questions evaluated the group repositioning behavior along the following dimensions: artificial, appealing, socially rich, disconnected, aware, appropriate, involved, and natural.

% For each video pair, we also included four questions using 5-point Likert scales to assess how welcoming each group appeared. At the end of each trial, participants were also given the opportunity to provide optional comments about the videos.

\section{Result and Analysis}
In this section, we first present and analyze the objective metrics from the ablation study, and then further examine these findings through the user study results and participant feedback.

\subsection{Objective Metrics}
\begin{table}[t]
\vspace{2mm}
  \centering

  \caption{\textbf{Quantitative analysis on ablation study.} Collision, jerk, and personal-space means cover eight setups; O-space covers seven eligible setups, excluding the two-agent elevator. E: ESFFM; M: MPC; C: CBF.}

  \label{tab: ablation}
  \resizebox{\columnwidth}{!}{%
  \begin{tabular}{lcccc}
    \toprule
    Controller  & \makecell{Mean collision\\rate (\%)} $\downarrow$ & \makecell{Mean jerk\\(m/s\textsuperscript{3})} $\downarrow$ & \makecell{Mean personal-\\space violation (\%)} $\downarrow$ & \makecell{Mean O-\\space violation (\%)} $\downarrow$\\
    \midrule
    SFFM       & 57.23 & 18.46 & \textbf{21.64} & 1.53 \\
    E      & 1.99 & 21.34 & 33.72 & 1.33 \\
    E+C  & 0.23 & 16.27 & 33.53 & 1.61 \\
    E+M                & 1.15 & 18.65 & 33.33 & 
    \textbf{0.74} \\
    E+M+C            & \textbf{0.18} & \textbf{14.63} & 32.83 & 1.76 \\
    \bottomrule
  \end{tabular}
  }
\end{table}

To examine the effects of MPC and CBF on robot-group repositioning behavior, we conducted an ablation study with five controller configurations: Pedica's SFFM, ESFFM, ESFFM+CBF, ESFFM+MPC, and ESFFM+MPC+CBF (ours)—across eight conditions, yielding 40 recorded trials. For each condition, the externally controlled robot was first teleoperated by one researcher, and its trajectory was recorded in a ROS 2 bag. The same recorded trajectory was then replayed for all five controller configurations within that condition to ensure a consistent external disturbance across models.
Each trial was evaluated from the first common odometry sample following the teleoperated robot's first nonzero velocity through the last common odometry frame before the recording ended. Motion smoothness is measured as the mean root-mean-square planar jerk across the autonomous robots. Collision rate denotes the percentage of evaluated frames containing at least one geometric collision. O-space violation rate denotes the percentage of evaluated frames in which at least one group's o-space is violated, while personal-space violation rate denotes the percentage of evaluated frames in which at least one robot's personal distance is violated (see Tab. \ref{tab: ablation}).

Because each condition was evaluated using a single prerecorded teleoperation trajectory, only one trial was available per controller configuration within each condition. The ablation results are therefore descriptive rather than inferential and are not used for statistical significance testing. Nevertheless, they remain informative for comparing controller behavior under identical external disturbances.

\textbf{\textit{Collision avoidance.}} Both CBF-based controllers achieved no detected collisions in 7 of 8 conditions, compared with 5 for ESFFM+MPC, 4 for ESFFM, and 1 for SFFM (see Tab. \ref{tab:collision}). The clearest benefit of CBF was observed in obstacle avoidance. Compared with the original SFFM, ESFFM incorporates environmental repulsion that allows robots to react to nearby obstacles. Nevertheless, because this avoidance emerges from local force interactions rather than trajectory planning or safety constraints, its effectiveness remains limited in complex or confined environments. Although our proposed framework substantially reduces collisions, it does not provide a global safety guarantee for the complete interaction. The teleoperated robot is neither controlled by the nominal controller nor corrected by the CBF, and therefore its motion can violate the safety assumptions imposed on the autonomous group members.

This setting also reflects an important characteristic of human--robot interaction. A human approaching, joining, or passing through a robot group is not governed by the robot controller and cannot be assumed to follow the same motion or safety rules as the robots. Thus, safety in such interactions is inherently shared: the robot group should reduce unsafe proxemic behavior, while the approaching human also retains responsibility for avoiding clearly unsafe motions.

\begin{table}[t]
\vspace{2mm}
  \centering
    \caption{\textbf{Non-zero collision rates (\%) by condition and collision type.} ``--'' indicates no detected collision. E: ESFFM; M: MPC; C: CBF.}
  \label{tab:collision}
  \footnotesize
  \resizebox{\columnwidth}{!}{%
  \begin{tabular}{lccccc}
    \toprule
    Condition & SFFM &E & E+M & E+C & E+M+C \\
    \midrule
    \multicolumn{6}{c}{Robot-Obstacle Collision} \\
    \midrule
    2 robots, elevator &31.98 & -- & -- & -- & --\\
    3 robots, elevator & 100.00 & 8.03 & 5.20 & --   & --   \\
    5 robots, hallway  & 59.74 & --   & 2.30 & --   & --   \\
    5 robots, crowded & 83.20 & -- & -- & -- & -- \\
    7 robots, crowded  & 100.00 & 3.44 & --   & -- & --   \\
    7 robots, hallway & 78.71 & -- & -- & -- & -- \\
    \midrule
    \multicolumn{6}{c}{Autonomous Robot-Teleoperated Robot Collision} \\
    \midrule
    5 robots, crowded & 2.10 & 1.56 & 1.72 & 1.87 & 1.45 \\
    5 robots, hallway & -- & 2.92 & --   & -- & --   \\
    7 robots, empty & 4.22 & -- & -- & -- & --\\
    7 robots, hallway & 7.10  & -- & -- & -- & --\\
    \bottomrule
  \end{tabular}
  }
\end{table}

\begin{figure}
    \centering
    \includegraphics[width=\linewidth]{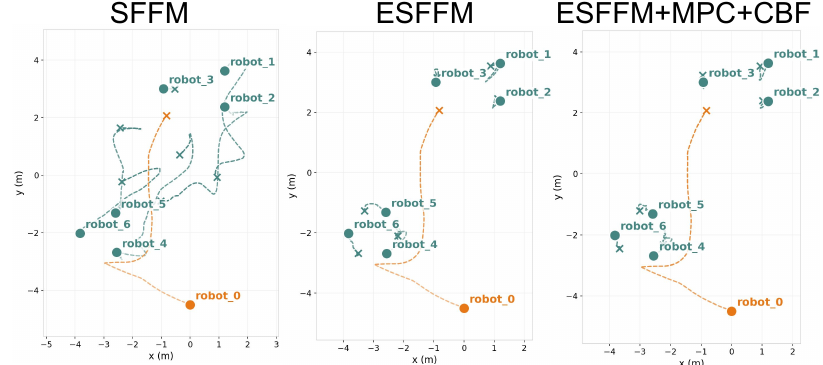}
    \caption{\textbf{Robot trajectories in the 7-robot crowded room layout.} The orange trajectory corresponds to the teleoperated robot, while the teal trajectories correspond to the 6 autonomous robots controlled by the respective controller. Circles (\textbullet) mark each robot's initial position and crosses (\texttt{x}) mark its final position.}
    \label{fig:trajectory}
\end{figure}

\textbf{\textit{Motion smoothness.}} Adding CBF to ESFFM+MPC reduced mean autonomous-robot RMS jerk in all conditions, with relative reductions ranging from $1.22\%$ to $39.63\%$. MPC without CBF also produced lower jerk than ESFFM without CBF in 6 of 8 conditions, except in empty space conditions. These results are consistent with the MPC objective’s acceleration-change penalty. For ESFFM, adding CBF reduced jerk in 5 conditions and increased it in 3. CBF therefore did not act as a universal smoothing mechanism. In 7-robot crowded room condition (see Fig. \ref{fig:trajectory}), the autonomous robots' trajectories became progressively smoother and more direct, indicating repositioning efficiency and reduced oscillatory behavior as the controller is augmented with MPC and CBF-based safety filter.

\textbf{\textit{O-space violations}}. O-space violation results were also condition-dependent. All controllers recorded 0 o-space violations in both empty space, the 7-robot hallway, and the 7-robot crowded room conditions. Meanwhile, the 3-robot elevator showed high violation rates. Adding CBF filter to the MPC controller also increased the o-space violation rate, and this is consistent with o-space maintenance remaining a nominal behavioral preference rather than an avoidance area constraint.

\textbf{\textit{Personal-space violations.}} Improvements in geometric collision avoidance did not consistently translate into better preservation of personal space. This is consistent with the difficulty of maintaining the specified interpersonal distances in confined environments such as elevators and hallways. The strong quadratic repulsive force in the original SFFM achieved the lowest personal-space violation rate. However, this came at the cost of less natural repositioning behavior, as group members moved faster and required longer reformation durations.

Overall, our proposed framework (ESFFM+MPC+CBF) provided the strongest observed combination of collision avoidance and motion smoothness, while social space preservation remained dependent on the environment and interaction geometry.

\subsection{Subjective Evaluation}
We evaluated how participants perceived the interaction quality of robot-group repositioning behaviors in the user study described in Sec. \ref{sec:user_study}. As mentioned before, ESFFM was used as the baseline instead of the original SFFM (having an extremely high collision rate) to provide a fairer comparison in environments containing obstacles.

\begin{figure*}[t]
\vspace{2mm}
    \centering
    \includegraphics[width= \linewidth]{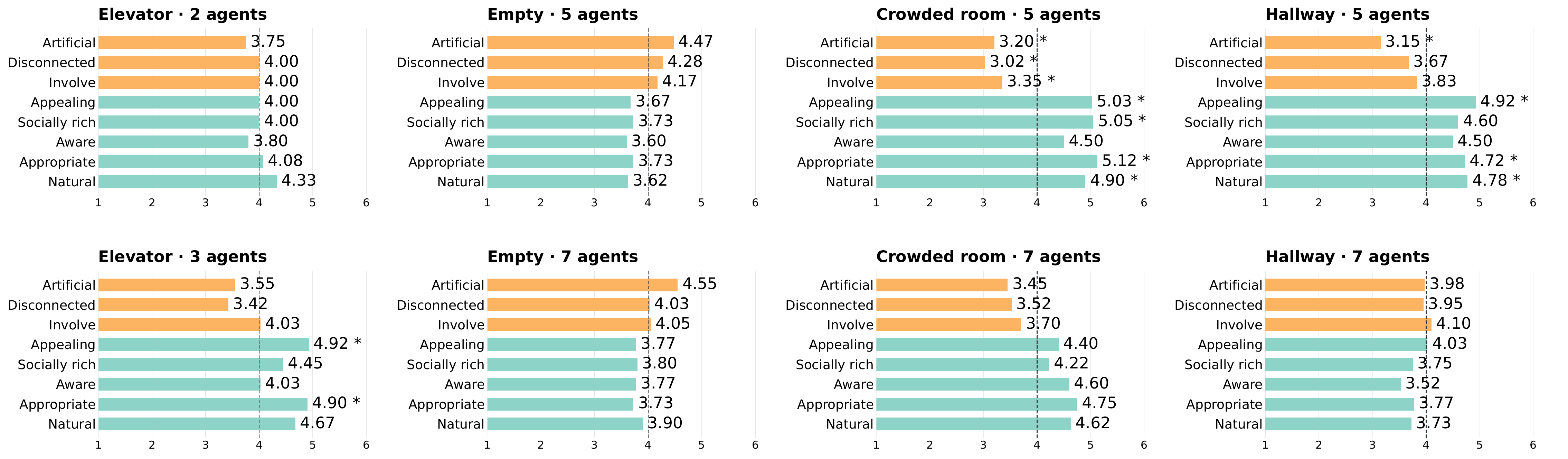}
    \caption{\textbf{Model preference by condition.} Ratings were made on a 7-point comparative scale with 4 indicating no preference. For negatively framed questions (orange), ratings below 4 favor our framework, whereas positively framed questions (teal), ratings above 4 favor our proposed framework (ESFFM+MPC+CBF). Within each condition, statistically significant deviations from the neutral midpoint after corrected one-sample \(t\)-tests are marked with ``*'' (\(p<0.05\)).}
    \label{fig:user_study}
\end{figure*}

We recruited 40 participants (22 men and 18 women; mean age = 34.4 years) through Prolific for the user study evaluation. Within each condition, we conducted two-sided one-sample t-tests on the 8 video-comparison rating questions with Holm correction (see Fig. \ref{fig:user_study}). Significant preferences for our proposed framework were observed in the 3-robot elevator condition for appealing and appropriate; in the 5-robot crowded room condition for artificial, appealing, socially rich, disconnected, appropriate, involved, and natural; and in the 5-robot hallway condition for artificial, appealing, appropriate, and natural (all $|t(39)| = 2.39\text{--}4.57$, Holm-adjusted $p = .0004\text{--}.043$). No ratings differed significantly after within-condition correction in the remaining conditions, and no rating significantly favored the ESFFM baseline in any condition.

% Separate linear mixed-effects models were fitted to the 8 comparison ratings with agent count, layout, their interaction, and scene order as fixed effects and participant as a random intercept. After Holm correction across 32 tests, layout remained significant for artificial, appealing, socially rich, disconnected, appropriate, and natural ratings ($\chi^2(2)=12.64\text{--}28.76$, adjusted $p\leq .049$). Crowded-room-versus-empty contrasts, corrected across 24 layout comparisons, favored our framework for all 6 measures (adjusted mean differences $=0.87\text{--}1.22$, $|z|=3.56\text{--}5.17$, adjusted $p\leq .007$). Agent count and its interaction with layout were not significant after correction, suggesting that the benefit was driven more by layout than by group size.

Separate linear mixed-effects models were fitted to the 8 comparison ratings with robot count (5 vs. 7), layout (empty, crowded room, hallway), their interaction, and scene order as fixed effects and participant as a random intercept. After Holm correction across all 32 effect tests, the effect of layout remained significant for artificial, appealing, socially rich, disconnected, appropriate, and natural; Wald $\chi^2(2) = 12.64 \text{--} 28.76$, with adjusted $p \leq .049$. We also directly compared the crowded room and empty conditions. After correction across all 24 layout comparisons, ratings favored our proposed framework in the crowded room for all 6 measures, absolute adjusted mean differences ranged from 0.87 to 1.22, and $|z| = 3.56 \text{--} 5.17$ (adjusted $p \leq .007$). Robot count and the robot-count-by-layout interaction did not remain significant after correction, suggesting that the relative benefit of our framework was associated with the crowded layout rather than with a general increase in the number of robots.

Although our proposed framework did not achieve the lowest social-space violation rates, participants still evaluated its repositioning behavior favorably, suggesting that improved safety and smoothness did not reduce perceived social appropriateness.

% Separate linear mixed-effects models were fitted to each of the eight comparison ratings, with the eight-level condition and scene order as fixed effects and participant as a random intercept. Likelihood-ratio tests with Holm correction across the eight questions showed significant condition effects for artificial, $\chi^2(7) = 32.98,~p < .001$; appealing, $\chi^2(7) = 28.72,~p < .001$; socially rich, $\chi^2(7) = 23.95,~p =.005$; disconnected, $\chi^2(7) = 22.27,~p =.007$; aware, $\chi^2(7) = 19.63, p =.013$; appropriate, $\chi^2(7) = 39.03, p < .001$; and natural, $\chi^2(7) = 30.14,~p < .001$. The condition effect was not significant for involved, $\chi^2(7) = 11.60,~p =.115$. These omnibus results indicate that at least one condition differed for each significant question. This shows that the complexity of the interaction condition affects how our proposed framework was evaluated against ESFFM.

% \section{Discussion}
% \input{ieeeconf/tex/discussion}

% \section{Future Work}
% \input{ieeeconf/tex/future_work}

\section{Conclusion}
% In this work, we presented a barrier-function-filtered receding-horizon controller for robot-group repositioning that improves interaction safety while preserving socially appropriate group behavior. The ablation study showed that the MPC component reduced motion jerk, while the CBF component further improved collision avoidance. The user study indicated a preference for our framework in perceived repositioning motion quality.
% The framework may also be applicable to virtual agents in video games and simulation platforms, or as a predictive model for trajectory planning in socially aware navigation. However, the current study is limited to simulated robots, and the social-force-based formulation still tends to produce symmetric group formations with limited naturalness and diversity. Future work will focus on physical-robot validation, and more flexible group formation behaviors.

In this work, we presented a barrier-function-filtered receding-horizon controller for robot-group repositioning that improves safety while preserving socially appropriate behavior. Ablation results showed that MPC reduced motion jerk and CBF further improved collision avoidance, while the user study favored our framework in perceived repositioning quality. The framework may also be applicable to virtual agents in video games and simulation platforms, or as a predictive model for trajectory planning in socially aware navigation. Limitations include simulation-only evaluation and symmetric social-force-based formations with limited diversity. Future work will focus on physical-robot validation and more flexible group formations.

% \section{Code}
% \textbf{Code availability.} The source code is maintained in a private GitHub repository pending future publication: \url{https://github.com/shay-zcx/sfm_cbf/tree/main}. I will ask the instructor for the TAs' email for access granting. But in case, if any member of the course staff is unable to access the repository and need to, please contact me at chuxuan\_zhang@sfu.ca and I will grant access immediately.

\bibliographystyle{ieeetr}
\bibliography{references}

\end{document}